# Automated detection of block falls in the north polar region of Mars

L. Fanara[1,2], K. Gwinner[1], E. Hauber[1] and J. Oberst[1,2]

[1] Institute of Planetary Research, German Aerospace Center (DLR)
[2] Geodesy and Geoinformation Science, Technical University Berlin

E-mail address: Lida.Fanara@dlr.de
Address: Rutherfordstr. 2, 12489, Berlin Germany

**Highlights:**

- A change detection method for block falls' identification on Mars is proposed.
- The images and a training data set are prepared.
- A combination of SVM/HOG and blob detection is used.
- The results show a true positive rate of ~75% and a false detection rate of ~8.5%.

## Abstract

We developed a change detection method for the identification of ice block falls using NASA's HiRISE images of the north polar scarps on Mars. Our method is based on a Support Vector Machine (SVM), trained using Histograms of Oriented Gradients (HOG), and on blob detection. The SVM detects potential new blocks between a set of images; the blob detection, then, confirms the identification of a block inside the area indicated by the SVM and derives the shape of the block. The results from the automatic analysis were compared with block statistics from visual inspection. We tested our method in 6 areas consisting of 1000×1000 pixels, where several hundreds of blocks were identified. The results for the given test areas produced a true positive rate of ~75% for blocks with sizes larger than 0.7 m (i.e., approx. 3 times the available ground pixel size) and a false discovery rate of ~8.5%. Using blob detection we also recover the size of each block within 3 pixels of their actual size.

**Keywords:** Change detection; block falls; machine learning; blob detection; Mars

## 3.1 Introduction

Steep scarps are prominent along the margins of the north polar ice cap of Mars (Fig. 1i, ii). Interestingly, images of these scarps acquired through the High Resolution Imaging Science Experiment (HiRISE McEwen et al., 2007) on the Mars Reconnaissance Orbiter spacecraft have revealed ongoing mass wasting activity, such as avalanches and ice block falls (Russell et al., 2012). Quantifying these block fall events over space and time will open a new window into the study of the scarps' dynamic evolution; however, this opportunity has yet to be taken advantage of due to the amount of time and resources necessary to undertake this task. The manual identification of newly emplaced, small, meter-sized blocks among the vast volume of available images, with each image covering tens of kilometres of scarp lengths, presents a challenging task. In order to circumvent this massive demand on human resources while yet taking advantage of all images, we have developed an automated method for block fall detection using comparisons of images taken at different times.

Existing change detection methods for Mars have been applied to large features such as impact craters, dark slope streaks and dust devil tracks (Di et al., 2014; Wagstaff et al., 2010; Xin et al., 2017). Techniques for rock detection have been developed mainly for autonomous rover navigation (Di et al., 2013; Gor et al., 2001; Thompson and Castano, 2007) and landing site characterisation, in particular for the site of the Phoenix (Golombek et al., 2008) and Mars Science Laboratory (MSL) spacecraft (Golombek et al., 2012). The algorithm relating to the latter detected rocks greater than 1.5 m in diameter based on shadow segmentation, which benefitted from the Phoenix landing site being flat and having a homogeneous, simple background (Golombek et al., 2008). The algorithm was improved to detect rocks greater than 1.2 m for the MSL landing site selection. However, the higher topographical complexity of the landing sites led to the requirement of support from a human operator in order to distinguish between shadows of rocks and other, non-rock shadows.

Blocks, similarly to rocks, belong to the smallest detectable features given the best available resolution of Mars' satellite imagery. However, with the north polar region being one of the most dynamic regions of the planet (Bourke et al., 2008; Russell et al., 2008, 2012), the aforementioned methods would struggle. In particular, the north polar regions of interest in our study have very steep topography casting various non-block shadows, while rough landslide deposits create a highly variable background in the images. Moreover, a pervading ice and dust layer can sometimes conceal the shadows, particularly of small blocks. Thus, relying on shadow segmentation for the detection would work well for the detection of large blocks but would introduce many false positive detections due to detecting non-block shadows and would often fail in detecting small blocks. These confounding factors mean that the detection of block falls at the north polar scarps of Mars requires a different approach.

Machine learning is key to automation in computer vision. The learning process extracts information (features) from a number of images depicting the object of interest (training dataset), with the selection of the appropriate features that best describe the object of interest being important for the success of the method. However, in cases where there exist a great number of training data (in the order of millions), this step can be performed by the machine learning algorithm itself (deep learning) with the results outperforming traditional computer vision methods (Krizhevsky et al., 2012). Unfortunately, such a number of training samples for ice block falls on Mars is not available to date. Therefore, this study utilises a classic combination of a machine learning classifier and a feature descriptor that became widespread after succeeding in pedestrian detection: a Support Vector Machine (Burges, 1998; Chang and Lin, 2011) with a Histogram of Oriented Gradients (HOG) descriptor (Dalal and Triggs, 2005). An advantage of the HOG descriptor is that it captures the object's local shape while being invariant to local geometric transformations as long as they are much smaller than the spatial sampling selected for the histograms (Dalal and Triggs, 2005). This makes the HOG very suitable for our task, where convex bright blocks of slightly varying shapes, commonly accompanied by a shadow in a specific orientation to the block, are to be detected.

We combine the SVM classifier with blob detection (Lindeberg, 1993, 1998), both for confirming the SVM detections and to extract the exact planar shape of each block, so that our method can serve for the monitoring of the north polar steep scarps and the estimation of their erosion rates.

## 3.2 Methods

### 3.2.1 Data

HiRISE images are very much suited for studies of block falls as the sun-synchronous orbit of the spacecraft and, consequently, repeated observations under very similar illumination conditions support change detection (McEwen et al., 2007). In particular, certain north polar scarps have been imaged over 100 times through the lifetime of Mars Reconnaissance Orbiter and thus multiple times per year, facilitating the identification of the short time interval within which these events occurred, potentially allowing us to study seasonal rates of block falls. However, the north polar region is covered by the seasonal $CO_2$ ice cap during winter that sublimates in the spring (Hansen et al., 2013; Piqueux et al., 2015). Hence, the visual appearance of the area is very different between the seasons.

We developed a method to compare images acquired during the summer seasons of consecutive years, ideally taken at similar solar longitude (i.e., time of the Martian year), when the environment appears most similar in the images. We trained and tested our method using two grayscale HiRISE images of the scarp at 83.80° N, 235.54° E taken one year apart: ESP_027750_2640_RED (from here on called the 'before' image) and ESP_036888_2640_RED (from here on called the 'after' image). These two images were taken with an angle between the spacecraft and a normal vector from the surface at the centre of the image (emission angle) of ~0° and an angle between the sun and the normal of the surface (incidence angle) of ~66° and ~70° respectively, providing very similar viewing conditions, as the majority of HiRISE images do. We selected six areas of 1000×1000 pixels for the validation of the method and collected the training dataset from the rest of the 'after' and the difference image's area.

We chose the validation areas so that they represent different scenarios that are common along the scarp (Fig. 1i). Areas a and d were already populated with a large number of blocks varying in size and shape, across several different layers close to the steepest active part of the scarp (up to 70° slope), the North Polar Layered Deposits (NPLD). Areas b, c and e have a more uniform and static background. The last area (Fig. 1f) is further downslope from the NPLD, with a large number of blocks appearing in the 'after' image, while the background of parts of the area differ clearly from the 'before' image. All six regions include parts with no activity in order for the method to be tested there as well.

We produced a Digital Terrain Model (DTM) and the ortho-recitfied images with the open-source NASA software Ames Stereo Pipeline (ASP) (Shean et al., 2016). ASP utilises the USGS Integrated Software for Imagers and Spectrometers (ISIS) tools for the preparation of the images described in the next section. We developed our method using the open source library OpenCV (Bradski, 2000).

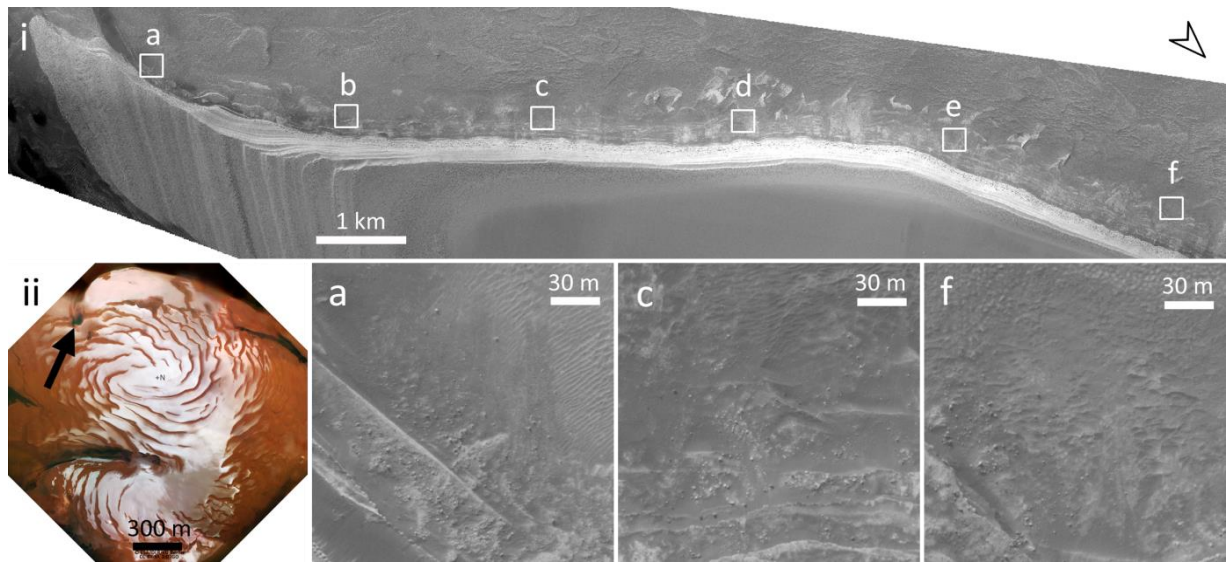

Figure 1. Steep scarp in the north polar region of Mars. i) part of HiRISE 'after' image ESP_036888_2640_RED, sun direction from top left, ii) High Resolution Stereo Camera (HRSC) colour mosaic of the north polar region of Mars (image credits ESA/DLR/FU Berlin) with arrow showing the location of the scarp in i, a, c, f) threes of the six validation areas.

### 3.2.2 Co-registration
The automated change detection requires two precisely co-registered images. To correct for the perspective distortions caused by the steep topography, images need to be ortho-rectified onto a sufficiently detailed DTM with a grid size comparable to the images' ground pixel size. Consequently, we can only investigate regions where at least one stereo pair of HiRISE images exists, from which a topographic model can be computed. We select stereo images which were taken under a stereo angle of 10°-25° and a time interval between the images of maximum 30 days. In this study we used the images ESP_019047_2640_RED and ESP_019298_2640_RED.

We use ASP to process the images, to produce a DTM and to ortho-rectify the images. First it processes the raw 10 CCDs in which HiRISE images are initially delivered with tools provided by ISIS. More specifically, it applies radiometric calibration correction and mosaics the 10 CCDs to a single HIRISE image. For the DTM production we choose a polar stereographic projection and a ground pixel size of 10 m. Then we ortho-rectify the images of interest using ASP. Some very small mis-registrations between the ortho-recitified images are compensated for by splitting the images into tiles of 200×200 pixels and performing an additional co-registration step. First, we prepare the image tiles by applying a bilateral filter to reduce the noise while preserving the boundaries of the objects, the so-called edges, and by normalising the range of intensity values of one image tile with respect to the other one. Then, the subpixel co-registration is realised by a simple translation, where translation parameters are found through Enhanced Correlation Coefficient Maximization (ECC) (Evangelidis and Psarakis, 2008).

### 3.2.3 Change detection

The co-registered image tiles are subjected to our change detection algorithm, which is a combination of a Support Vector Machine (Burges, 1998; Chang and Lin, 2011) with blob detection (Lindeberg, 1998, 1993) (Fig. 2), with the former identifying the regions of potential new blocks and the latter confirming the existence of a block and deriving its shape.

We subtract the 'before' from the 'after' image, creating a difference image. In this difference image, the newly appeared blocks are distinguished from other changes as they are bright convex objects, usually accompanied by darker regions, their shadows. Although we benefit from the specific identical relative orientations of blocks and their shadows both in the original images and the difference image, we also detect blocks that have no strong accompanying shadow.

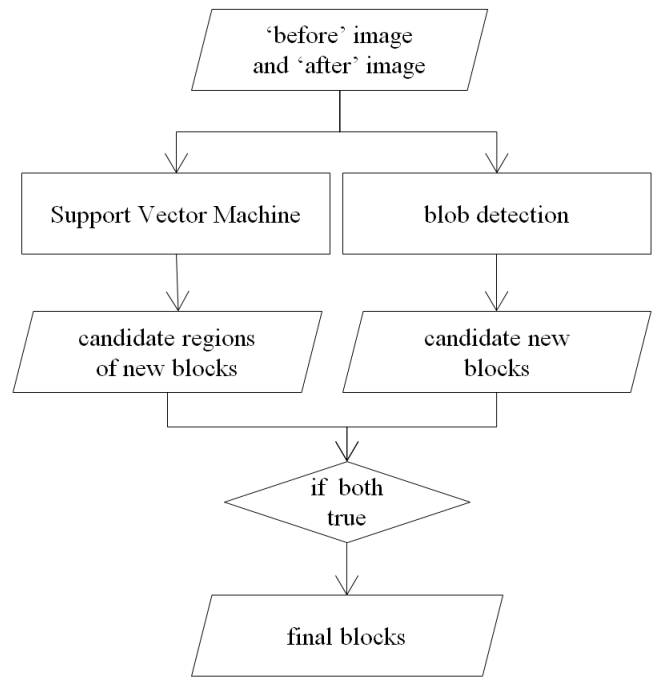

Figure 2. Flowchart of change detection method. Trapeze: input/output, rectangle: process, diamond: decision.

**3.2.3.1 Support Vector Machine**

We trained an SVM on the 'after' image and another SVM on the difference image based on Histograms of Oriented Gradients (HOG) (Dalal and Triggs, 2005), avoiding the six areas that we selected for validating the method.

For training the SVM we collected thousands of examples of images showing blocks (positive samples) as well as images not containing any block (negative samples). More specifically we selected 3,750 positive and negative samples from the 'after' image and 3,028 positive and negative samples from the difference image. The positive samples were cropped parts of the image each containing a block located in its top half and the shadow in the bottom half (Fig. 3a). The negative samples were cropped parts of the image, containing other features found in the image, with a focus on those resembling blocks (Fig. 3b). The positive samples were cropped with a side ratio of 4:5, which was suitable for their shape when accompanied by a shadow and easy to be resized to 64×80 pixels to conform to the publication that presented HOG (Dalal and Triggs, 2005). We extracted the HOG descriptor (Fig. 4) of each block as well as of multiple 64×80 rectangles from each negative sample and used the resulting set of HOG descriptors to train an ε-Support Vector Regression linear SVM with ε=0.1 and C=0.01.

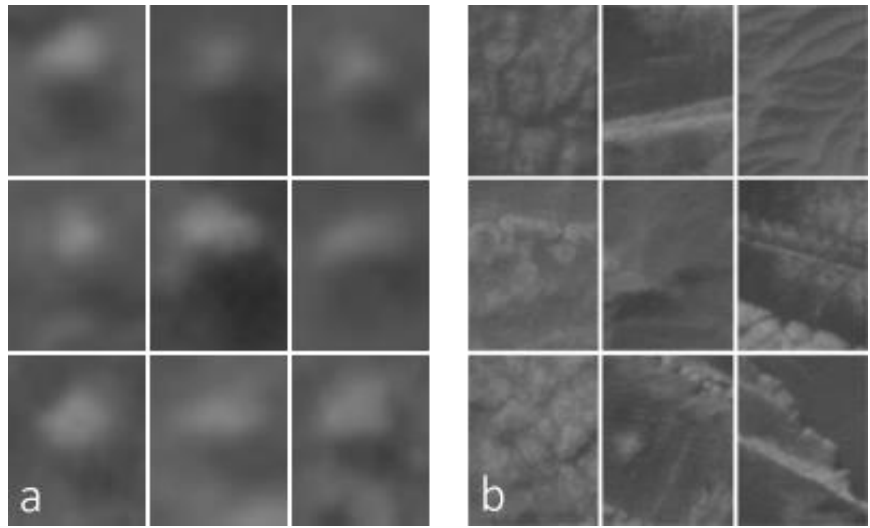

Figure 3. Training dataset examples, sun direction from top left. a) positive samples from the 'after' image (resized to 64×80 pixels), b) negative samples from the 'after' image.

We detect regions of candidate blocks in the difference image based on the support vector computed from training on the difference image and, similarly, in the original images based on the support vector computed from training on the 'after' image. We first enlarge the images 8 times in order for the smallest detectable blocks (∼ 8×10 pixels including the potential shadow and some background) to be detectable using a window of 64×80 pixels. This is the smallest reasonable window to compute a HOG descriptor with the recommended binning of 8×8 pixels to find the shape of a block (Fig. 4). We then perform multi-scale detection, with a detection window increase factor of 1.05 and a hit-threshold of 0.5. This results in 3 maps of candidate regions of blocks in the 'before', 'after' and difference images (Fig. 5).

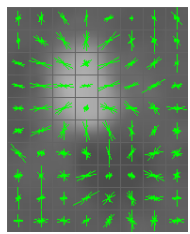

Figure 4. HOG descriptor of a resized to 64×80 pixels block.

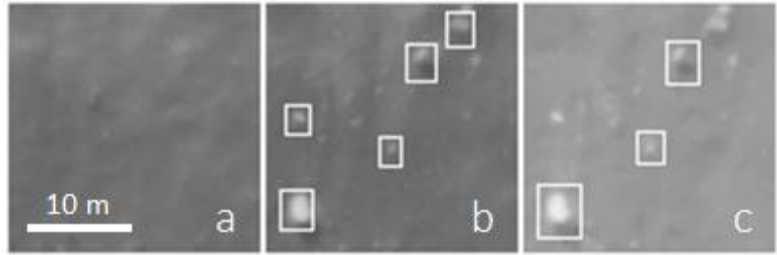

Figure 5. SVM detections, sun direction from top left. a) candidate block regions on the 'before' image, b) candidate blocks' regions on the 'after' image, c) candidate blocks' regions on the difference image.

### 3.2.3.2 Blob detection

To confirm the existence of a new block in an area indicated by the SVM-HOG detections and to determine its shape, we combine simple blob detection and Maximally Stable Extremal Regions (MSER) (Nistér and Stewénius, 2008) with Canny edge detection (Canny, 1986) (Fig. 6).

We first highlight the areas that changed between the two images (Fig. 7a, b) by applying thresholding to the difference image (Fig. 7c) (intensity: 0-255) to derive the dark (here <123) and bright areas (here >133) (Fig. 7d). The thresholds originate from histogram analysis of the difference image and represent half a standard deviation of the mean intensity value. The MSER detection and the simple blob detection acquire bright and dark regions of the difference image that differ from their surroundings based on these same thresholds (Fig. 7f, g respectively). The Canny edge detector extracts the edges of the difference image based on the intensity gradients of neighbouring pixels, using two thresholds, one for the minimum gradient of strong edges and one for the minimum gradient of weak edges. We set them to 30 and 15 respectively based on simple tests that we run on the difference image, looking for a combination that derives the boundaries of both large and small blocks without picking up other weak edges found inside large blocks (Fig. 7e).

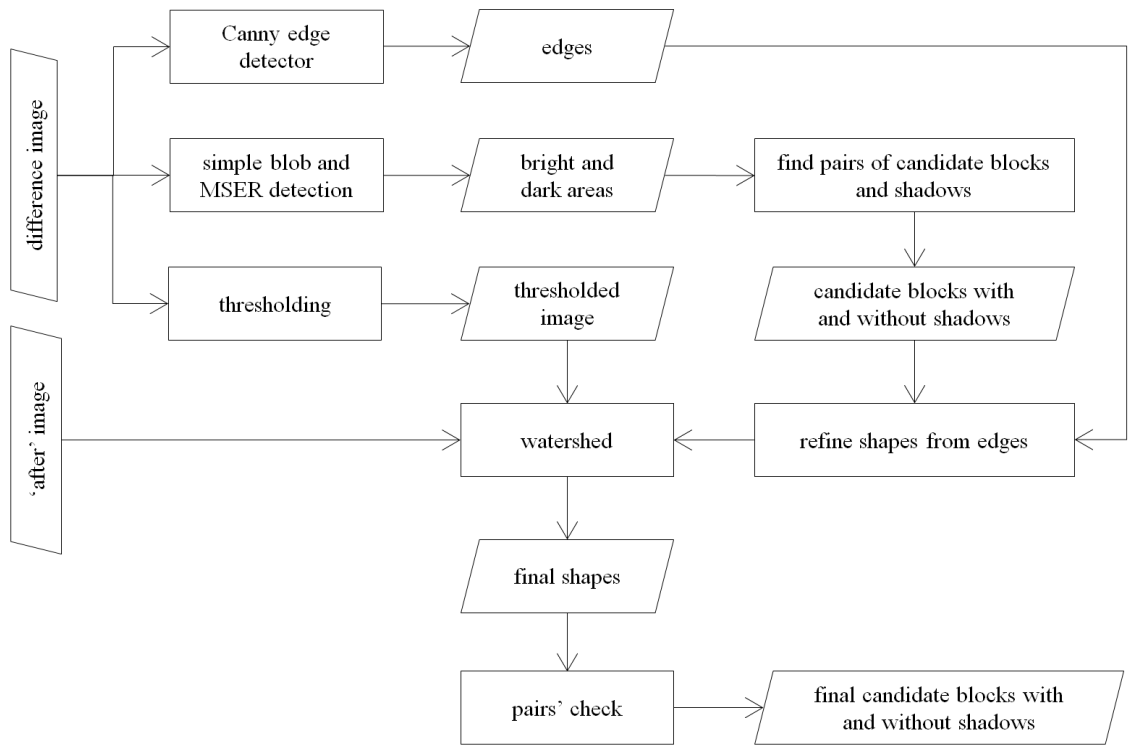

Figure 6. Flowchart of blob detection. Trapeze: input/output, rectangle: process.

We search for potential block-shadow pairs amongst the bright and dark areas by checking if there is a shadow close to a block and approximately in the opposite direction of the sun (Fig. 7h). We consider bright areas without an adequate shadow as blocks only if both the simple blob detection and MSER identified this area. We use the detected edges to refine the shape of the candidate blocks (Fig. 7i) before applying the watershed algorithm (Meyer, 1992), a region growing algorithm, using as markers the exclusive disjunction between the threshold image and the refined shapes to finalise the shapes of blocks and shadows based on the 'after' image (Fig. 7j). Given the final shapes of blocks and shadows, we check if they still form pairs with the right orientation to the sun or if they fulfil the requirements for being accepted as shadow-less blocks (Fig. 7k). The result of the blob detection is a map of all final candidate blocks (Fig. 7l).

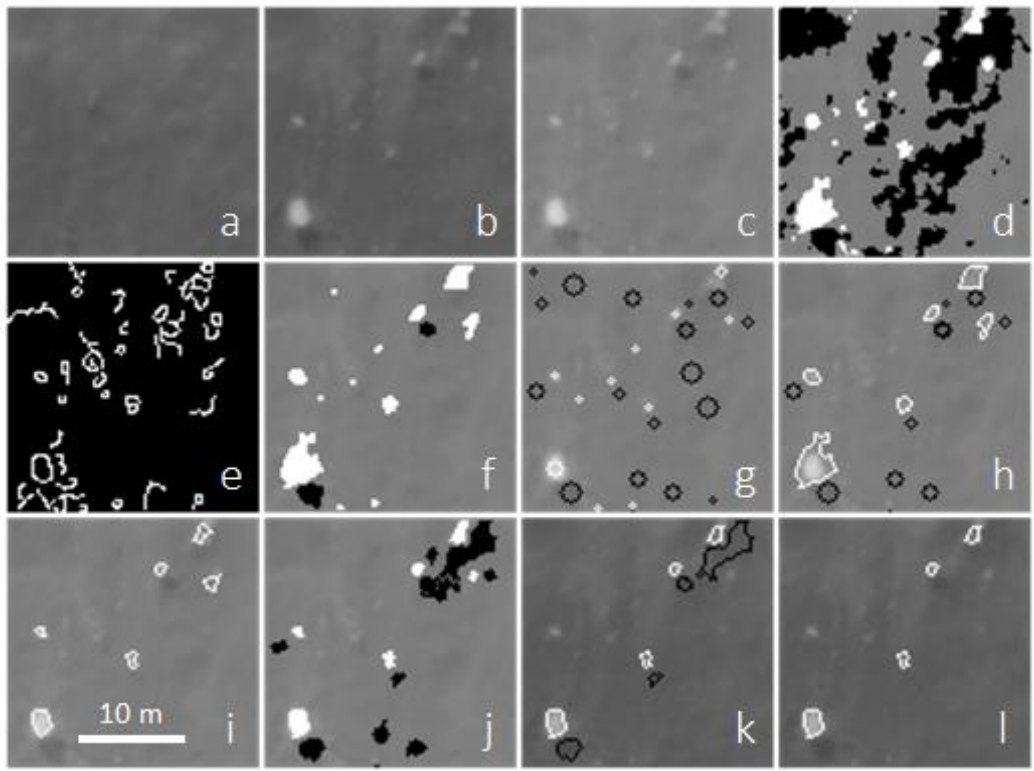

Figure 7. Blob detection, sun direction from top left. a) 'before' image, b) 'after' image, c) difference image, d) threshold image, e) Canny edge detection, f) MSER bight and dark areas, g) simple blob bright and dark areas, h) candidate blocks and shadows, i) updated shape of candidate blocks from edges, j) shape of candidate blocks and shadows after watershed algorithm, k) final pairs of blocks and shadows, l) final candidate blocks.

### 3.2.3.3 Final new block detection

If a candidate block with a shadow detected by the blob detection is found within an area indicated by the trained SVM in the 'after' or the difference image and no detection occurred at the same location in the 'before' image, then it is accepted as a newly appeared block. A candidate block without a shadow must have been detected by the SVM both in the 'after' and in the difference image to be accepted as a new block. The result of our algorithm is a map of the final fresh blocks.

## 3.3 Results and discussion

We evaluated the results from our automatic detection by comparing them to statistics from the manual identification of fresh blocks in six areas characterised by different backgrounds and block abundances, each representing a different case that can be encountered along the north polar scarps (Fig. 8). We assessed our method on its ability to detect the blocks and estimate their individual sizes.

### 3.3.1 Block detection

We found a True Positive Rate (TPR) of 61.18% and a False Discovery Rate (FDR) of 10.57% for all new blocks appearing in the validation areas (Table 1). However, as expected, the ground pixel size of 0.25 m limited the detection of objects with an area smaller than ~0.5 $m^2$ (i.e. with a diameter less than 3 times the pixel size). We, therefore, analysed the performance of the method for blocks with an actual area greater than 0.5 $m^2$ separately, and the TPR improved to 75.07% with the FDR decreasing to 8.50%. Although the ground pixel size of the images prevented the identification of the majority of blocks smaller than 0.5 $m^2$, our method detected a significant portion of them (37.84%). In spite of the complexity of the background, characteristic of the environment of the north polar scarps of Mars, the number of false detections (FDR = 8.50%) was low, which highlighted that our method can be reliably used automatically without supervision.

As outlined in Table 1, our method performed best in the area with the most uniform and static background (Fig. 8c), achieving a TPR of 87.50%. The most active region (Fig. 8f), where a large event led to some blocks appearing very close to each other and a changing background, had a TPR of 67.62%. At the same time, however, the false positives were the lowest with an FDR of 1.39%. In the validation areas with the most complex backgrounds (Fig. 8a, d), which represent the most common case along the scarp, our TPR was 77.97% and 75.53% respectively. This is a high value for such a varied terrain and with the achieved low FDR, our method is a powerful tool for fast unsupervised monitoring of steep north polar scarps' activity.

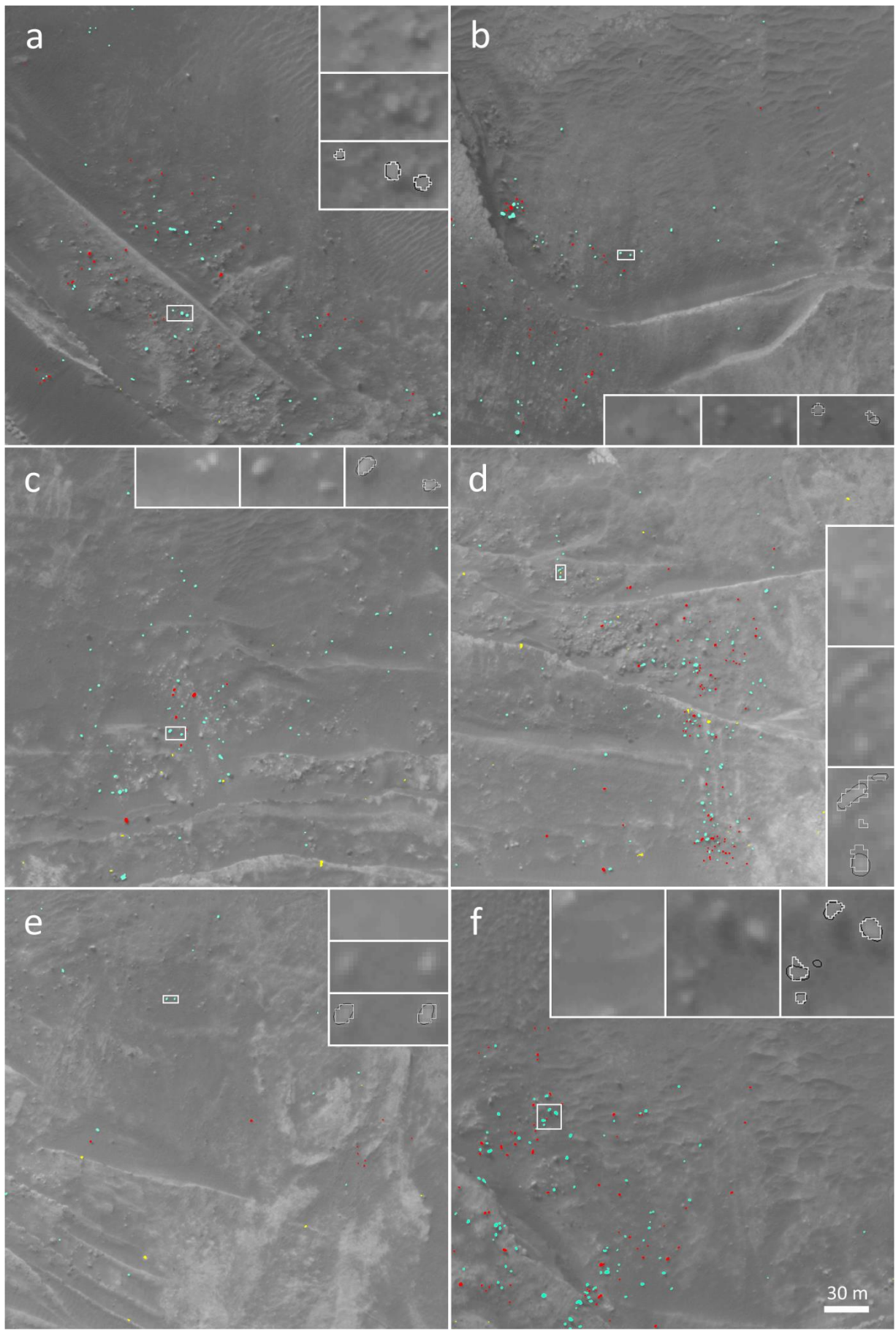

Figure 8. Validation results for automatically detected new blocks in six different areas (a-f). Note the true positives in turquoise, the false negatives in red and the false positives in yellow. The inset of each region shows the region in the white box in 'before' and 'after' images, note the actual blocks (black) and the detections (white).

Table 1. True Positive Rate and False Discovery Rate.

|  |  | total actual | total predicted | true positives | True Positive Rate (%) | False Discovery Rate (%) |
|---|---|---|---|---|---|---|
| region a | all | 100 | 61 | 59 | 59.00 | 3.28 |
|  | > 0.5 m² | 59 | 50 | 46 | 77.97 | 8.00 |
| region b | all | 80 | 48 | 47 | 58.75 | 2.08 |
|  | > 0.5 m² | 48 | 35 | 34 | 70.83 | 2.86 |
| region c | all | 71 | 73 | 63 | 88.73 | 13.70 |
|  | > 0.5 m² | 56 | 56 | 49 | 87.50 | 12.50 |
| region d | all | 180 | 122 | 101 | 56.11 | 17.21 |
|  | > 0.5 m² | 94 | 79 | 71 | 75.53 | 10.13 |
| region e | all | 24 | 21 | 15 | 62.50 | 28.57 |
|  | > 0.5 m² | 11 | 14 | 9 | 81.82 | 35.71 |
| region f | all | 140 | 82 | 79 | 56.43 | 3.66 |
|  | > 0.5 m² | 105 | 72 | 71 | 67.62 | 1.39 |
| total | all | 595 | 407 | 364 | 61.18 | 10.57 |
|  | > 0.5 m² | 373 | 306 | 280 | 75.07 | 8.50 |

### 3.3.2 Block area estimation

Besides identifying new blocks, our method estimated their area, and we analysed its performance (Fig. 9). In general, it derived the planar area of the blocks within 3 pixels, similar to the accuracy reported by Golombek et al. (2012) using HiRISE data (1-2 pixels in diameter). The fitted straight line to the data has a slope of 0.8 and an offset of 0.1, implying that the sizes of the blocks were estimated accurately. The average error was 3.2 pixels. However, our method has the tendency towards underestimating the area of large blocks, which was probably caused by their shadows obscuring parts of the block. Overall, the results presented here demonstrate that our automatic technique can be reliably used to estimate total volumes of widely spread block populations from block falls, and e.g., minimum erosion rates on scarps quickly and effectively, which represents a massive improvement over the currently used visual inspection, thus fulfilling the objective set out at the beginning of this study.

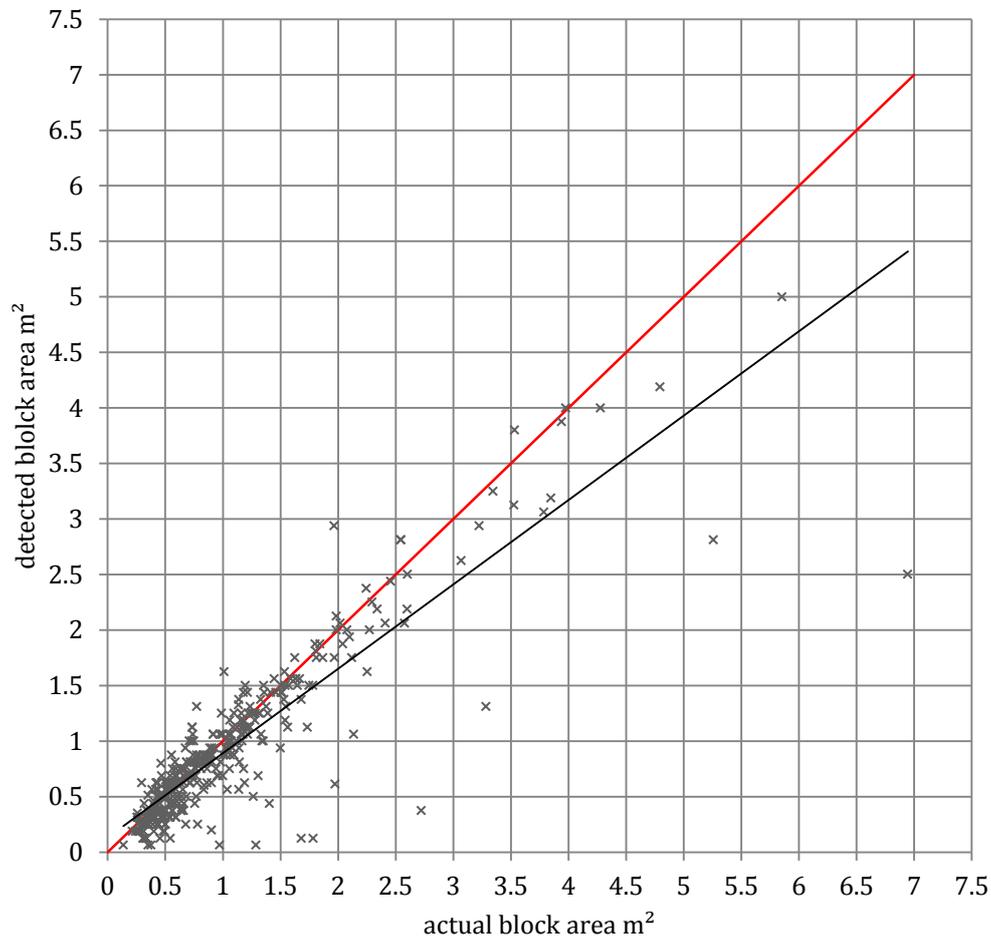

Figure 9. Performance in detecting the area of each block with straight black line fitted to the data ($R^2 = 0.8$), average error is 3.2 pixels.

## 3.4 Conclusions

This study presents the first method designed for the automated detection of block falls in NASA's HiRISE images of the north polar scarps on Mars, with these events representing the smallest detectable changes on the surface of the planet. For blocks larger than 0.5 m² (i.e., with a diameter exceeding the image pixel size by a factor of 3), the true positive rate of the method is 75.07% and the false discovery rate 8.50%. Importantly, we demonstrate that the method can be applied without manual intervention. Therefore, it can effectively be utilised to monitor all north polar scarps using images spanning the lifetime of HiRISE in order to produce a comprehensive map of block fall events. Resulting event statistics can facilitate the estimation of the erosion rate of the scarps, a finding which in turn would vastly improve our understanding of the evolution of the ice cap closely linked with the recent Martian climate. In the future, we will use machine learning for detecting block falls in images produced in different seasons at the north polar scarps and to identify blocks in other parts of the planet. Further extending our method to be trained to detect and map other surface changes over large regions of Mars, will inform us about the rates of present-day geological activity on a dynamic planet.

**Acknowledgements**

This research has made use of the USGS Integrated Software for Imagers and Spectrometers (ISIS). This work was co-funded by the European Union's Seventh Framework Programme under iMars grant agreement n° 607379.